\documentclass[journal]{IEEEtran}
\usepackage{xcolor,soul,framed}
\usepackage[pdftex]{graphicx}
\graphicspath{{../pdf/}{../jpeg/}}
\DeclareGraphicsExtensions{.pdf,.jpeg,.png}
\usepackage{array}
\usepackage{mdwmath}
\usepackage{mdwtab}
\usepackage{eqparbox}
\usepackage{url}
\usepackage{float} 
\usepackage{subfigure}
\usepackage{gensymb}
\usepackage{amsmath}

\usepackage[T1]{fontenc}
\usepackage[utf8]{inputenc}
\usepackage{authblk}

\begin{document}

\title{Dandelion-Picking Legged Robot}
\author[1]{Sandilya Sai Garimella\thanks{garimell@umich.edu}}
\author[2]{Shai Revzen\thanks{shrevzen@umich.edu}}
\affil[1]{Department of Mechanical Engineering, University of Michigan, Ann Arbor}
\affil[2]{Department of Electrical Engineering and Computer Science, Department of Ecology and Evolutionary Biology, University of Michigan, Ann Arbor}

\renewcommand\Authands{ and }
\maketitle

\begin{abstract}
Agriculture is currently undergoing a robotics revolution, but robots using wheeled or treads suffer from known disadvantages: they are unable to move over rubble and steep or loose ground, and they trample continuous strips of land thereby reducing the viable crop area.
Legged robots offer an alternative, but existing commercial legged robots are complex, expensive, and hard to maintain.
We propose the use of multi-legged robots using low-degree-of-freedom (low-DoF) legs and demonstrate our approach with a lawn pest control task: picking dandelions using our inexpensive and easy to fabricate BigANT robot.
For this task we added an RGB-D camera to the robot.
We also rigidly attached a flower picking appendage to the robot chassis.
Thanks to the versatility of legs, the robot could be programmed to perform a ``swooping'' motion that allowed this 0-DoF appendage to pluck the flowers.
Our results suggest that robots with six or more low-DoF legs may hit a sweet-spot for legged robots designed for agricultural applications by providing enough mobility, stability, and low complexity.
\end{abstract}

\begin{IEEEkeywords}
agricultural robotics, mechanical implement, robotic picking, multi-legged robots, mobile manipulation, computer vision
\end{IEEEkeywords}

\section{introduction}



\IEEEPARstart{T}{he} conventional approach to harvesting agricultural produce, such as fruit, is to use large harvesting machines to produce mechanical vibration to separate the fruit from the stalk or to use human labor for manual picking.
Both these methods have downsides: the former method fails to reap a third of edible produce harvested (33.7\% market yield remains on the farms) \cite{BAKER2019541} and there is no guarantee that the plants or picked fruit remain intact due to damage from machinery \cite{BAKER2019541}.
The latter method requires trained pickers and involves health hazards such as lacerations \cite{mccurdy2003agricultural}.
Existing wheeled harvesters such as the Octinion Rubion and the strawberry-picking system from Dogtooth Technologies require produce to be cultivated in tailored environments like raised beds \cite{bogue2020fruit} and they cannot be used in a field.
Other wheeled robotic pickers depend on sophisticated mechanical design, for example, the apple picker from FFRobotics operates between 4 and 12 robotic arms and requires human supervision \cite{bogue2020fruit} and the wheeled raspberry harvesting robot from Field Robotics relies on 4 robotic arms \cite{bogue2020fruit}. Purchasing and maintaining several agricultural machines and robotic pickers is expensive, requires training, and they are not easily transportable, thus dissuading farmers with small farms (defined by the USDA as farms with gross cash income under \$250,000 annually) from using them \cite{akram2020study}.

Any machine designed to support agriculture must either be the size of the agricultural fields themselves or be able move through them.
Conventionally, such mobility employs large wheels or treads to push against the ground -- thereby regularly trampling a substantial fraction of the potential growing area.
Additionally, wheels and treads encounter trouble when the ground is soft or flowing, and cannot move over very sharp inclines or discontinuous terrain (like stepping stones).

Robotic platforms with legs might be a means to avoiding these disadvantages. 
Unfortunately, these typically have 3 or more DoF per leg, making even a quadruped require a minimum of 12 motors.
Here we provide an alternative to such complex multi-legged robots using BigANT, a 6-legged (hexapedal) robot with only one DoF per leg. 
Thus, BigANT has only half as many motors as the typical quadruped robot, yet is far easier to stabilize even when running at speed - because it is statically stable.
The superior value proposition of multi-legged robots with low-DoF per leg was recently presented in \cite{zhao2020multi, Zhao-2021-PhD}, where the author studied the design, modeling, and control of such robots. 
This class of robots offer the advantages of legged systems without the complexity and cost associated with having numerous actuators.

As a model problem, we considered the task of removing dandelions from a lawn.
Dandelions are a known lawn pest, and they reproduce extremely quickly. 
Because their bright yellow flowers are easy to detect on the green background of a lawn, problems of machine vision and target acquisition are minimized.
We addressed the dandelion-picking problem by generating a series of actions from a library of existing behaviors (walking, steering, and turning in place), some of which are parametric.
We chose which behavior to perform based on the RGB-D data which we first reduced to the dandelion's azimuth, elevation, and distance.
Below we provide background on the BigANT robot and its behaviors (section \ref{sec:prev}), followed by an investigation of the techniques used to cut the stem of a dandelion (section \ref{sec:cut}).

\begin{figure}[!htb]
  \begin{center}
  \includegraphics[width=2.5in]{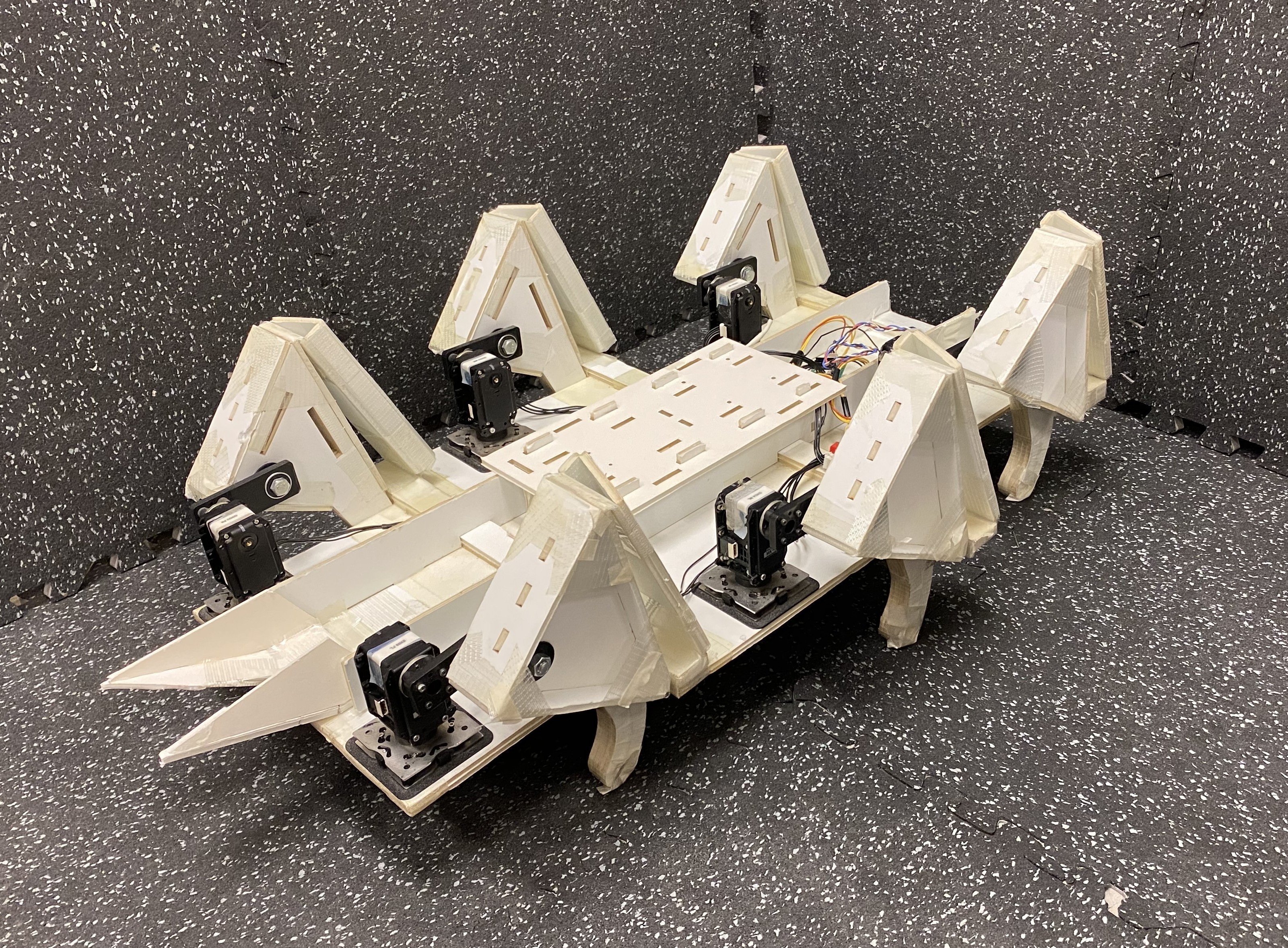}
  \caption{BigANT with cutting appendage attached to the front of the chassis.}
  \end{center}\label{fig:robot}
\end{figure}

\subsection{Previous work on the BigANT hexapedal robot}\label{sec:prev}
\begin{figure}[!htb]
    \centering
    \includegraphics[width=3.3in]{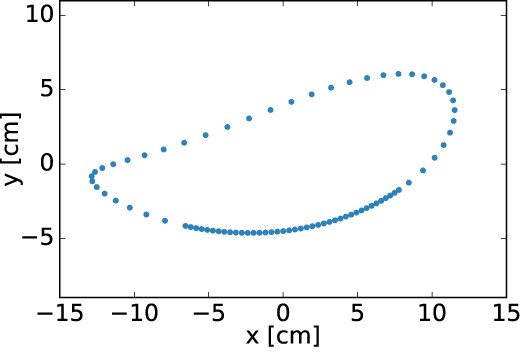}
    \caption{Toe trajectory in body frame, with points at equal phase intervals.
    Each leg pair's rate modulation and position in the cycle will be used to define a function for the chassis trajectory.
    Adapted from \cite{zhao2020multi}.}
    \label{fig:traj}
\end{figure}

This project was done entirely on the BigANT robot (Figure \ref{fig:robot}) which has six legs, each with 1-DoF.
Its properties are described in detail in \cite{zhao2020multi}.
To walk the robot we drove the six legs in an ``alternating tripod gait'', with the left tripod comprising of [F]ront-[L]eft, [M]iddle-[R]ight, and [H]ind-[L]eft legs (FL-MR-HL), and the right tripod being FR-ML-HR \cite{zhao2020multi}.
The toe trajectory (Figure \ref{fig:traj}) occupies a one-dimensional manifold with respect to the body frame of reference and choosing where the leg pairs (F, M, and H pairs) should be on this trajectory formed the basis of the overall BigANT chassis `swooping' motion which we developed to pick the dandelions.
The idea of exploiting the existing motors to produce the swooping motion came from work on limiting parasitic vertical chassis oscillations when designing the tripod gait -- where Zhao exploited the timing of leg motions to reduce the vertical motions.
By doing (mostly) the opposite, we exploited the same individual leg trajectories to lower and raise the robot while pitching, and produce the `swooping' motion to pick a dandelion.

By modifying the gait parameters (defined in \cite{zhao2020multi}) of the mid left and mid right legs we obtained a steering gait, e.g. slowing down the mid right leg and speeding up the mid left leg during the ground contact part of the trajectory results in steering right.
We varied a dimensionless quantity called the turn value $T$, which lies in the range $-0.3\le T \le 0.3$, to produce steering with a particular turn radius. 

\subsection{Methods for cutting a dandelion stem: chopping or slicing?}\label{sec:cut}
The ways in which a dandelion stem is cut can be distilled to three main categories: separation due to tensile fracture, applied normal force due to a blade (chopping), and applied normal and shear forces (slicing). 
For tensile fracture to occur, pulling along the stem axis of the dandelion is sufficient. 
During chopping the stem undergoes localized normal compression until failure occurs. 
Slicing and chopping are dissimilar because the former involves both normal and shearing deformations whereas the latter involves normal deformations only, causing global deformations \cite{reyssat2012slicing}. 
A soft solid such as a dandelion stem offered greater resistance to failure under compressive stress than it did to tensile stress so the slicing technique was more effective due to localized deformation.
Considering which cutting action to use was crucial because to produce any of the three cutting actions, the cutting device will need to have at least 1 DoF, either innately or by using BigANT's DoFs.
We explored the combination of two separation techniques, i.e. tensile stress with either chopping or slicing by manually testing cutting devices which we fabricated using rapid prototyping.

We tested the chopping idea by rapidly prototyping a box-like enclosure with a utility knife blade attached to the closing boundary of one of the parts. 
Once the dandelion stem was cut, it remained engulfed by the box, which was closed. 
The chopping action occurred when the blade pressed against the closing boundary of the opposite piece, with the dandelion stem trapped in between; the closing piece with the knife blade was actuated by a servo motor.

\begin{figure}[!htb]
    \centering
    \subfigure[]{
    \label{fig:desA}
    \includegraphics[width=1.8in]{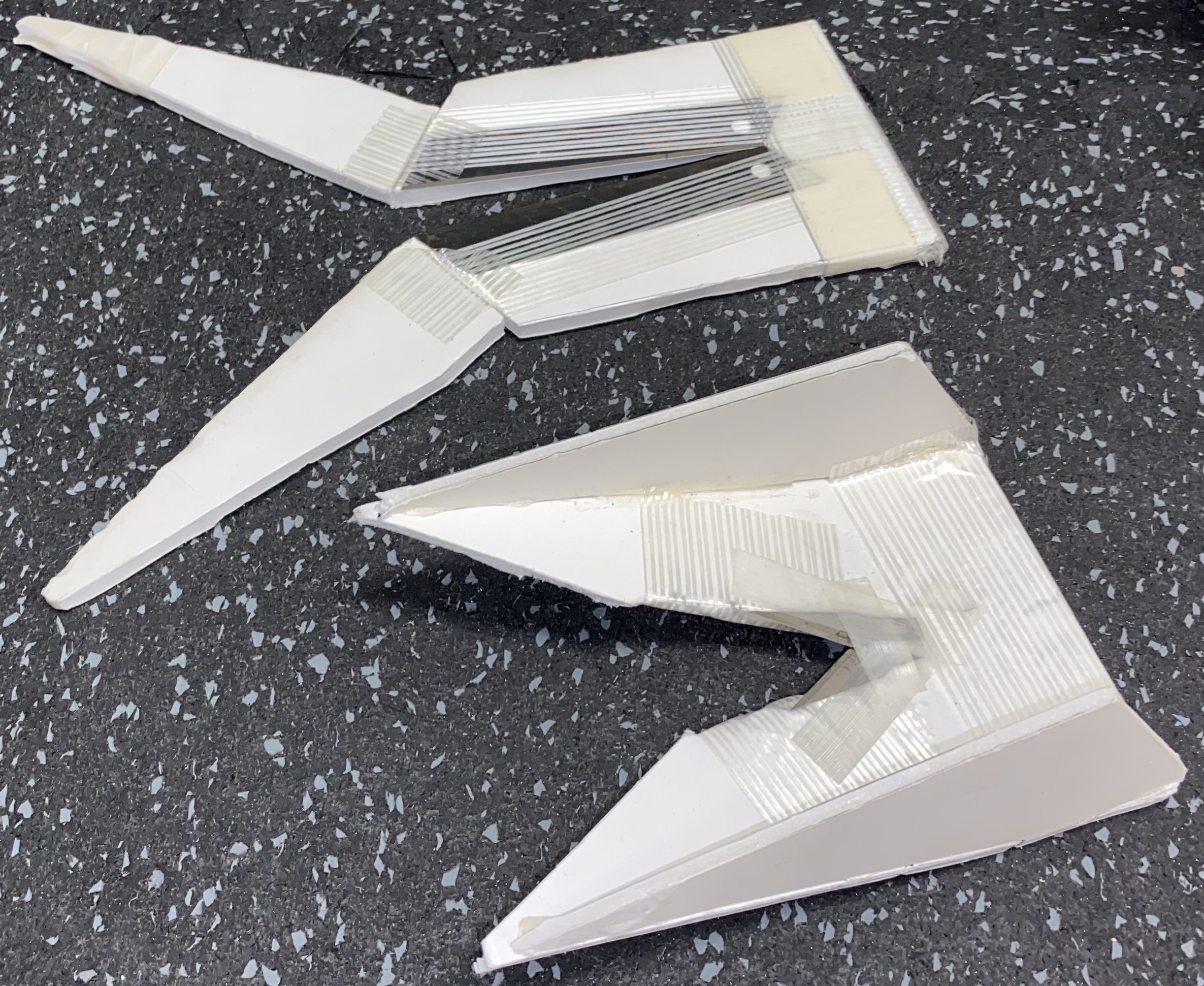}
    }
    \subfigure[]{
    \label{fig:desB}
    \includegraphics[width=1.4in]{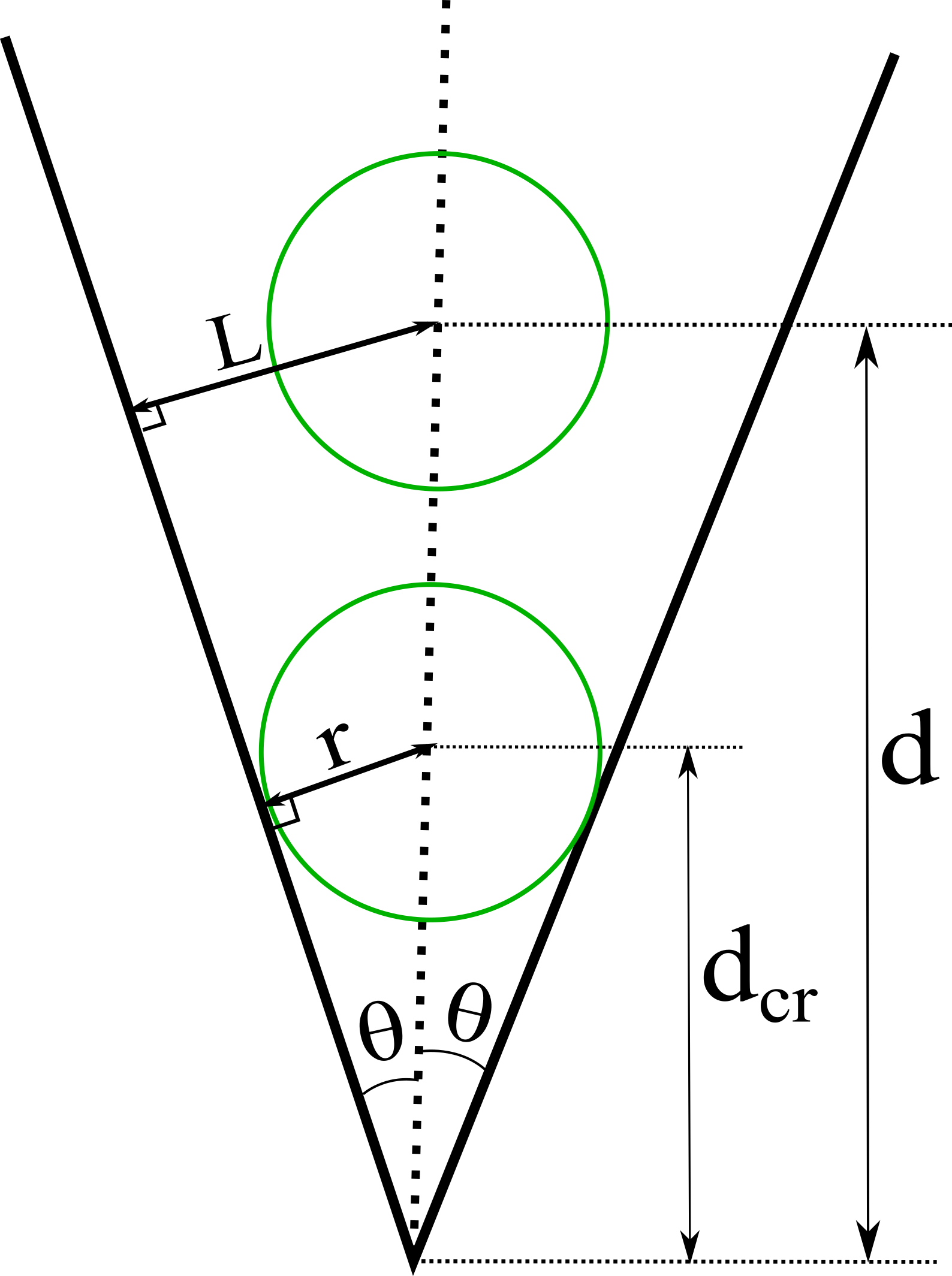}
    }
    \caption{(a) Comparison of the first cutting appendage (top) and second design.
    Note the improvements in the second iteration: positive inclination and side barriers for consistent plucking.
    In (b), the top view of `V' shape cutting appendage with a dandelion stem of radius $r$ approaching the vertex.}
    \label{fig:des}
\end{figure}

\begin{figure}[!htb]
    \centering
    \subfigure[]{
     \label{fig:outdoorA}
    \includegraphics[width=1.50in]{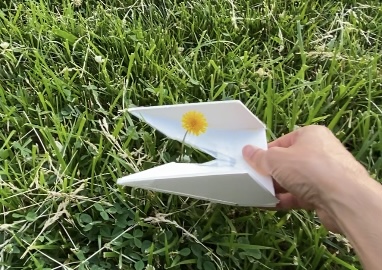}
    }
    \subfigure[]{
     \label{fig:outdoorB}
    \includegraphics[width=1.70in]{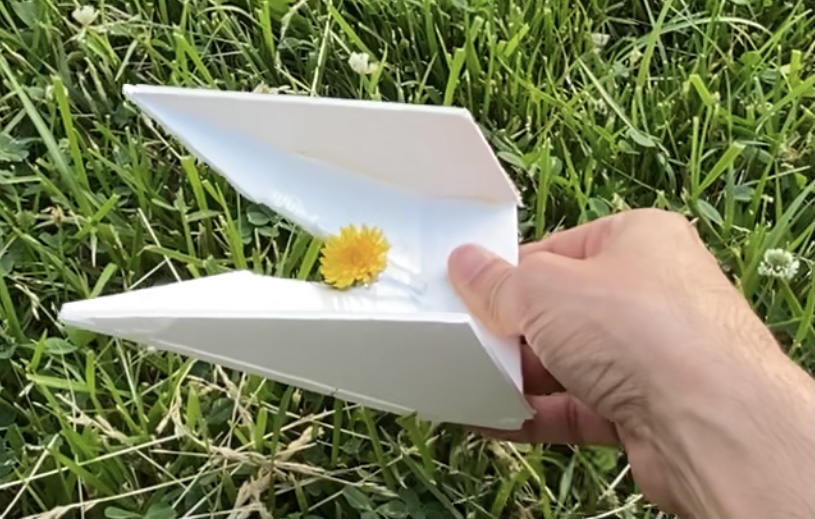}
    }
    \subfigure[]{
    \label{fig:outdoorC}
    \includegraphics[width=1.60in]{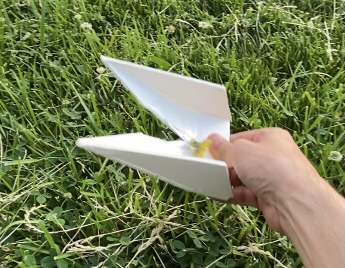}
    }
    \caption{Approaching the dandelion in early swoop trajectory in (a).
    The dandelion head is caught firmly at the vertex in (b).
    Due to tensile fracture, the dandelion is cut in (c).}
    \label{fig:outdoor}
\end{figure}

To test the slicing idea, we fabricated a `V' shaped cutting appendage using two utility knife blades fixed to a foam core frame. Rather than predicting the performance theoretically, we examined the design by slicing dandelions, and made improvements through two design iterations (Figure \ref{fig:desA}).
The testing method was to move the cutting appendage through a concave-up trajectory by hand (Figures\ref{fig:outdoorA}, \ref{fig:outdoorB}, and \ref{fig:outdoorC}) to simulate the BigANT producing the same motion via its chassis `swooping' trajectory.
Through our trials, we identified that dandelions were being sliced when the cutting appendage approached them at azimuthal angles near its center (near dashed line in Figure \ref{fig:desB}). By widening the two opening sides of the `V' shape, we increased the azimuth range in which the dandelion was picked successfully.
This increased the error allowance for the swooping trajectory and dandelion position to intersect after steering towards the dandelion.
With repeated use, we improved the first design's flaws in the second iteration, namely the lack of restraint for the dandelion after it is plucked and the cut dandelion occasionally falling forwards.
The second cutting appendage had foam core barriers to prevent the cut dandelion from falling sideways out of the platform, and a positive inclination with respect to the BigANT's chassis base to make it fall backwards consistently (Figure \ref{fig:outdoorC}).

Although the `V' shape cutting appendage has 0-DoF on its own, it harnesses the BigANT's forward motion to produce normal and shear stresses required for slicing.
As the dandelion stem approaches the vertex of the `V', the maximum stem radius which the blades can accommodate decreases.
The relationship between the maximum radius $L$, distance of the stem center to the vertex $d$, and half the interior angle of the `V' $\theta$, is given by $L = d\sin(\theta)$. The condition for slicing is satisfied when the stem center is at critical distance $d_{cr}$ from the vertex; then $d_{cr} = r/\sin(\theta)$, where $r$ is the stem radius (shown in green in Figure \ref{fig:desB}).

Depending on the height of the dandelion, either the tensile fracture or slicing occurred.
Suppose the dandelion height was greater than that of the cutting appendage blades above ground, slicing occurred because only the stem interacted with the blades.
If the dandelion was short, its head was positioned above the blades near the vertex until the final stage of the BigANT's `swooping' trajectory when enough tensile stress was applied for separation to occur.
This phenomenon is demonstrated in figures \ref{fig:outdoorB} and \ref{fig:outdoorC}. 

\section{A chassis fixture for picking dandelions}

The chosen cutting appendage would have to be move along an appropriate trajectory to operate.
One option was to attach it to an actuated mechanism designed to produce the needed motion.
Instead, we explored a solution that attached the cutting appendage to the chassis.
The existing DoFs which the cutting appendage could have due to BigANT's mobility were forwards and backwards movement (on BigANT's sagittal plane), yaw due to steering or turning in place, roll due to height differences on left and right legs, pitch through differences in height of front and hind legs, and vertical motion withing the clearance height of the legs.

\begin{figure}[!htb]
    \centering
    \includegraphics[width=3.3in]{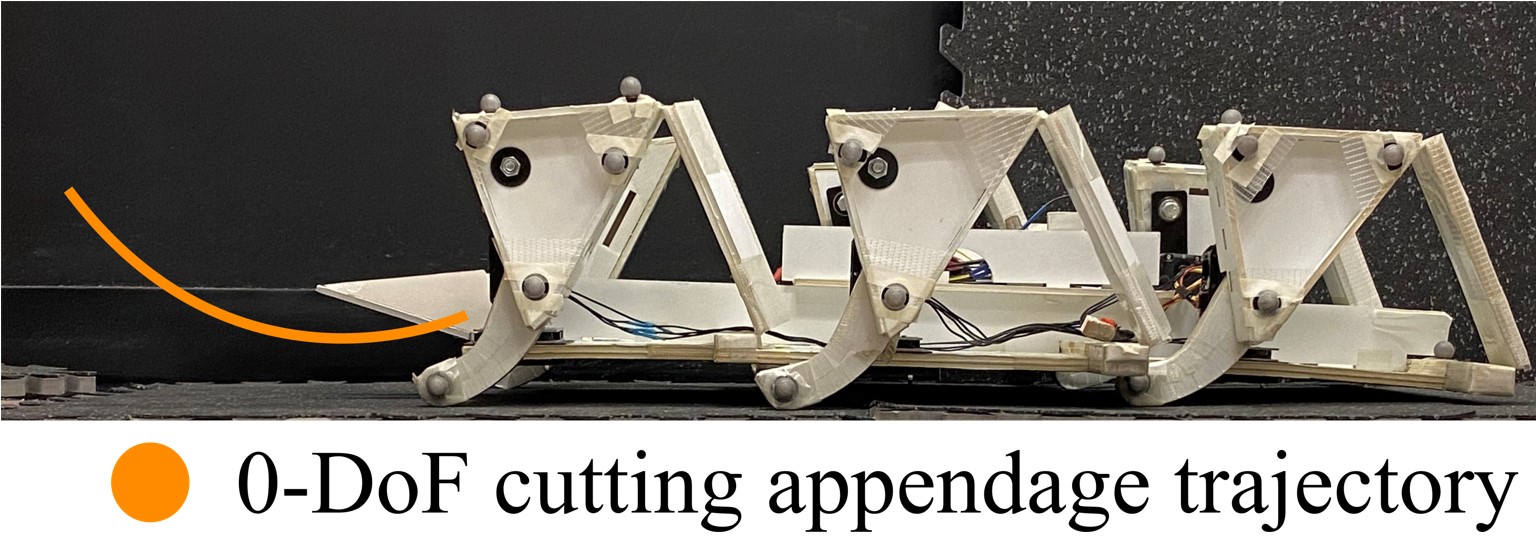}
    \caption{Swooping motion of the cutting appendage. The feasible toe trajectories relative to the current body location are shown in Figure \ref{fig:traj}. %
    We combined toe motions to produce the cutting appendage motion.}
    \label{fig:orange}
\end{figure}

By performing identical symmetric motions on right and left contralateral leg pairs, we reduced the complexity of the trajectory design from 3-D to 2-D.
From here on, in describing the picking maneuver, we will discuss it as if it is a 2-D problem.
The `swooping' motion we designed consists of lowing the cutting appendage, then moving it forward and up, while intercepting the dandelion near the nadir of the `swooping' arc.
We created motions (e.g. the cutting appendage trajectory in Figure \ref{fig:orange}) by using the rapid prototyping abilities developed in \cite{revzen2010ckbot}: puppeting the robot legs by hand, recording the motor positions in a gait table in CSV format, and replaying the that table as a motion primitive (a \texttt{Plan} in the language of \cite{revzen2010ckbot}).

\begin{figure}[!htb]
    \centering
    \subfigure[]{
    \label{fig:swoopA}
    \includegraphics[width=1.6in]{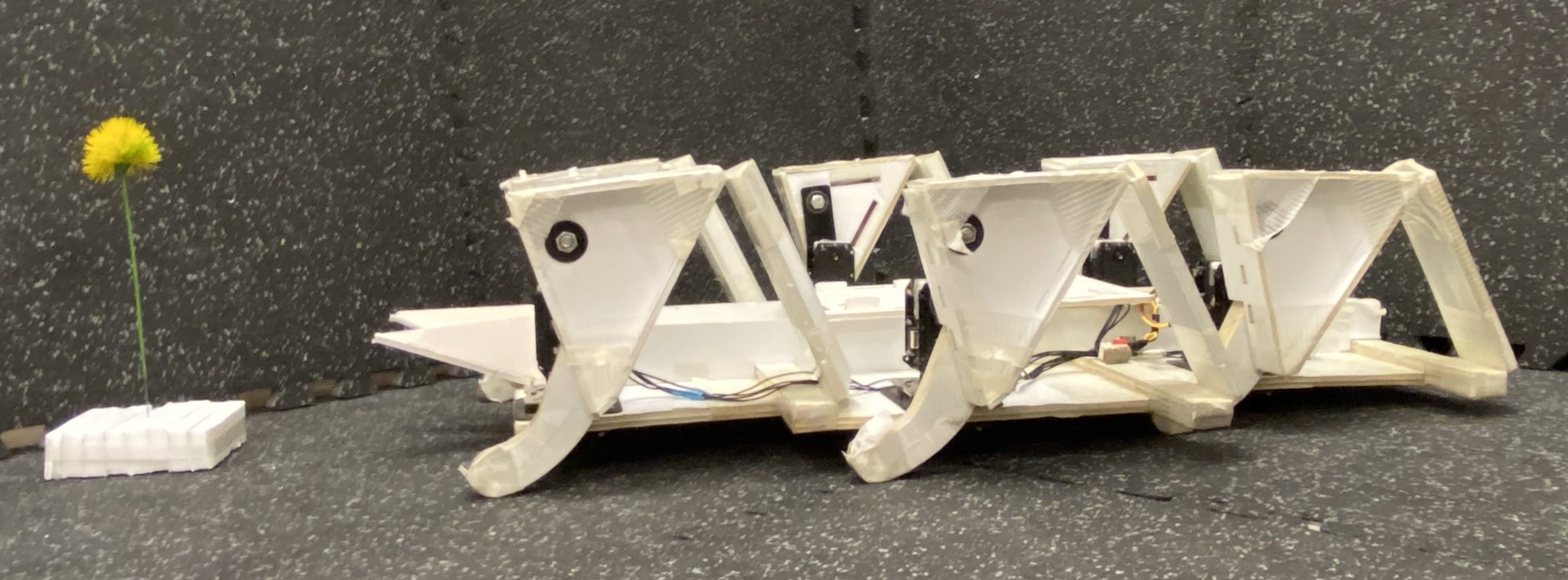}
    }
    \subfigure[]{
    \label{fig:swoopB}
    \includegraphics[width=1.6in]{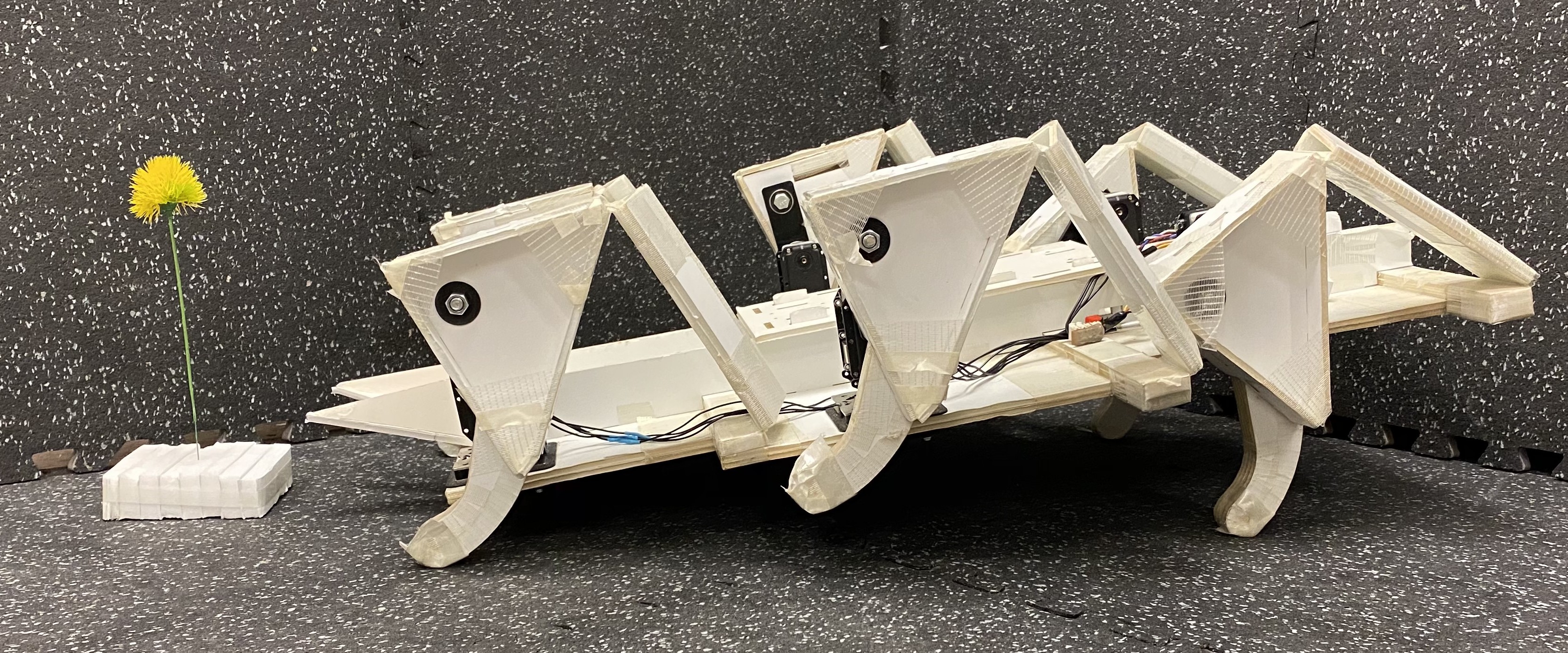}
    }
    \subfigure[]{
    \label{fig:swoopC}
    \includegraphics[width=1.6in]{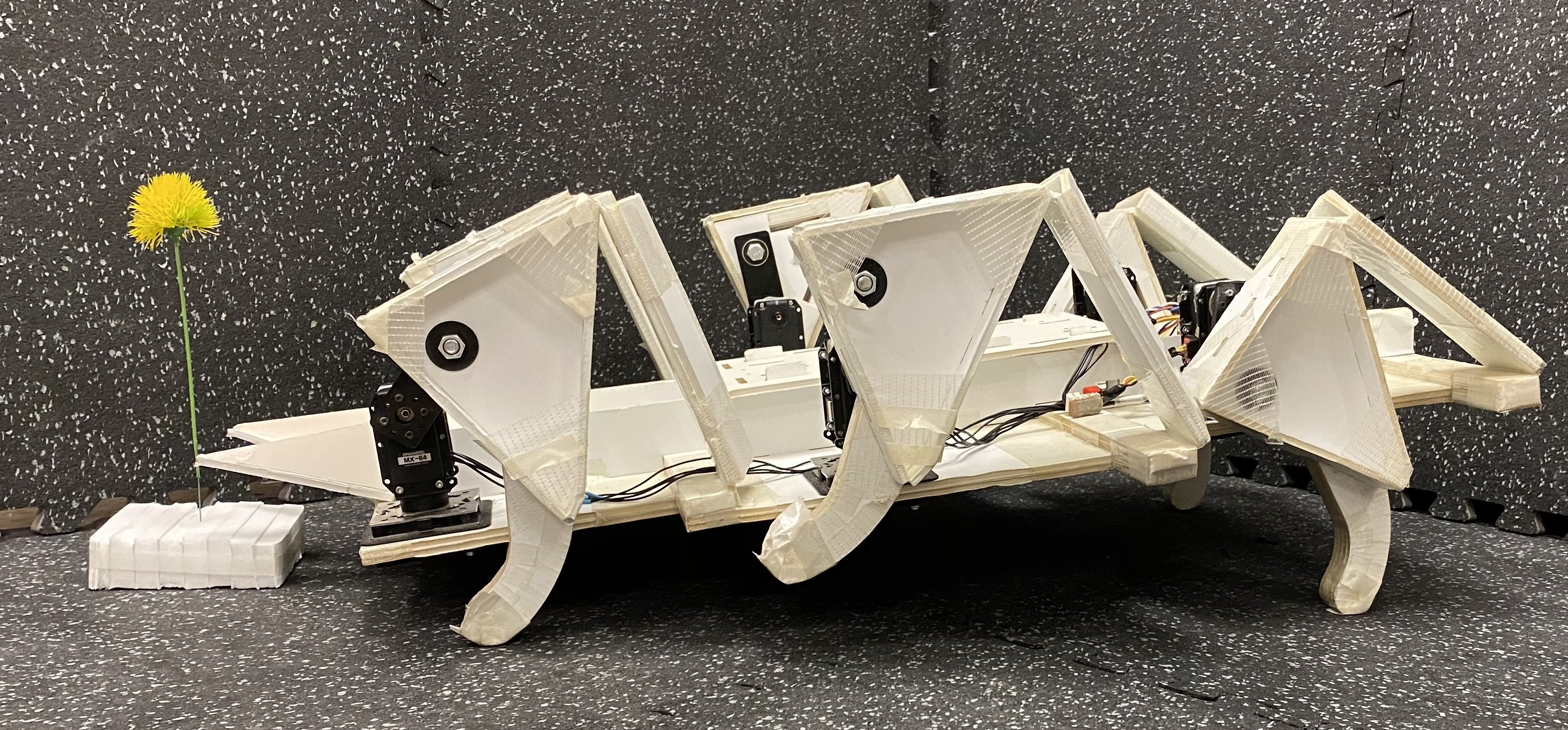}
    }
    \subfigure[]{
    \label{fig:swoopD}
    \includegraphics[width=1.6in]{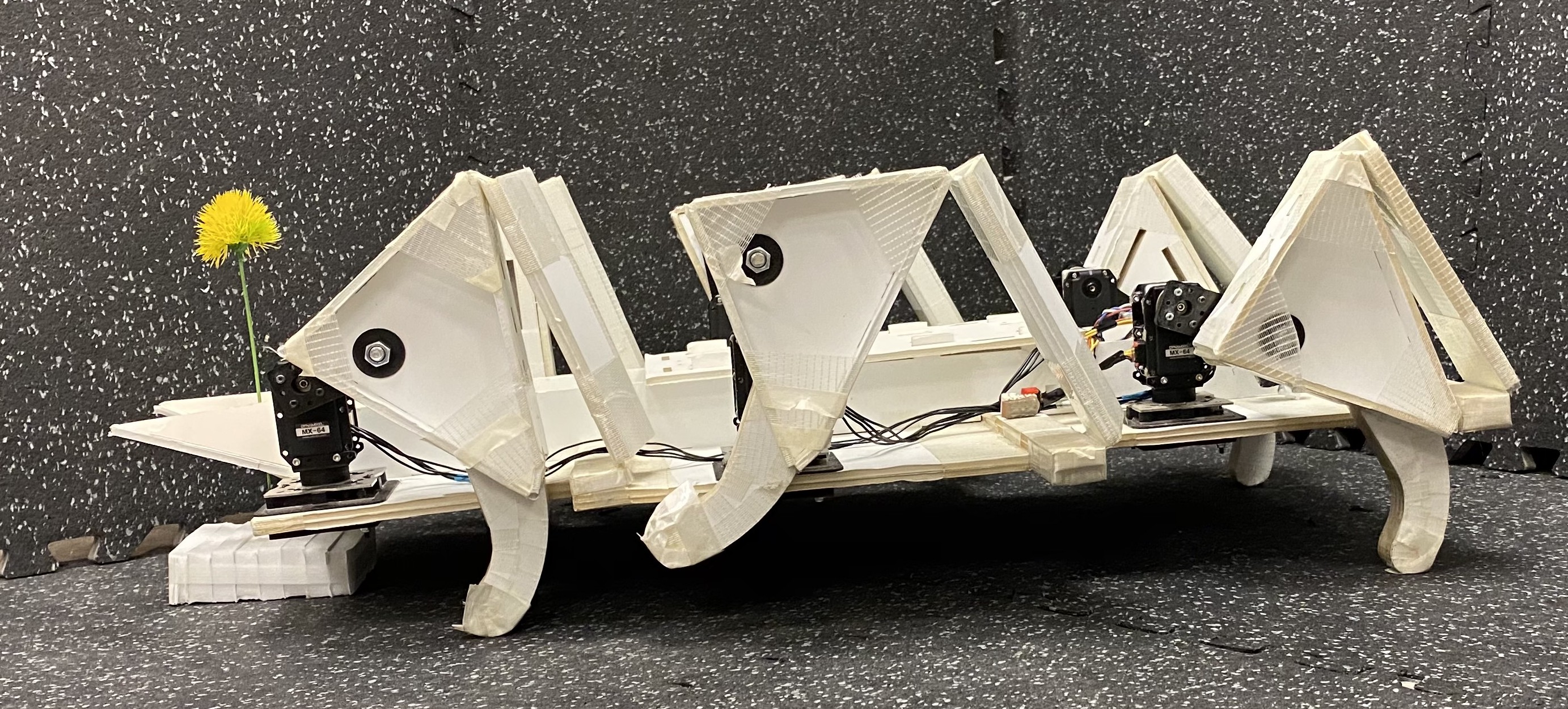}
    }
    \subfigure[]{
    \label{fig:swoopE}
    \includegraphics[width=1.6in]{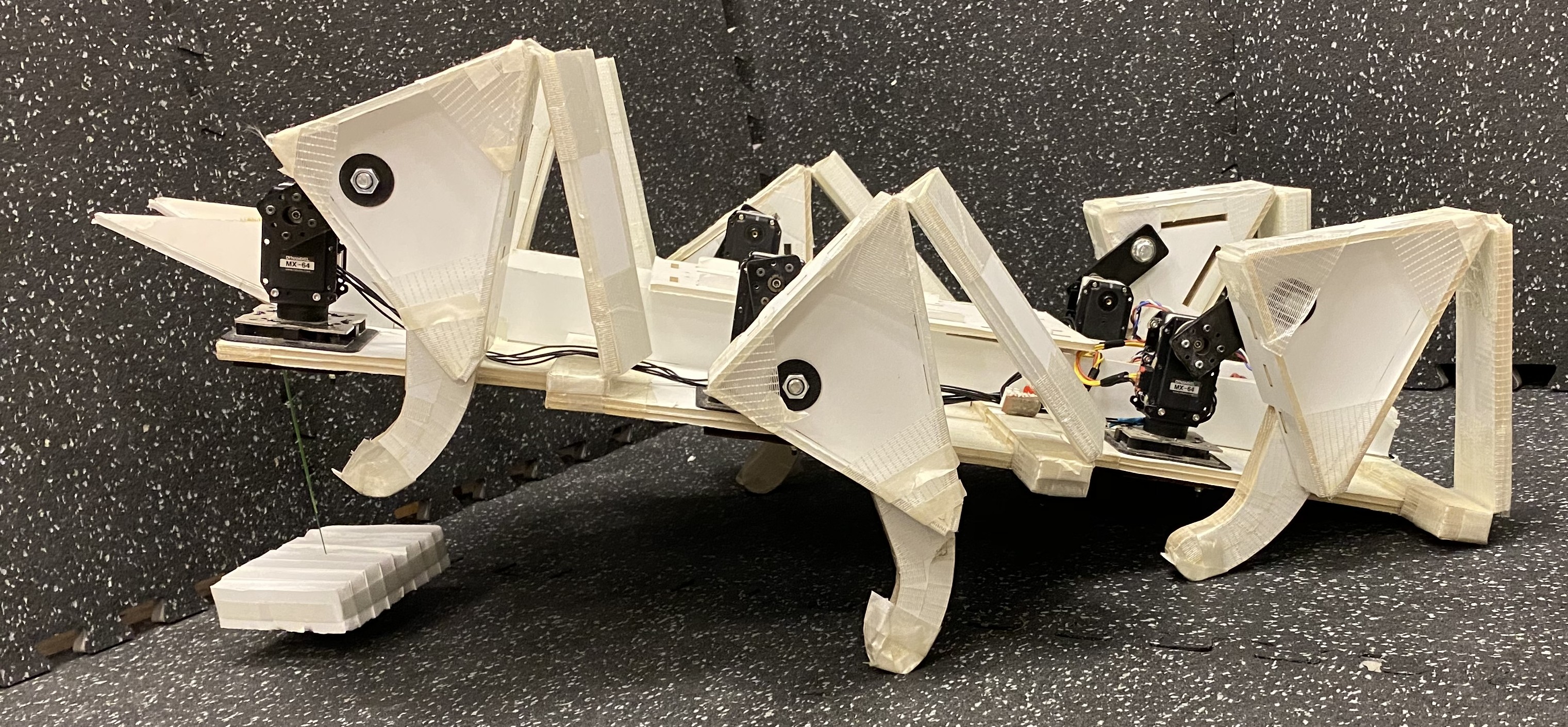}
    }
    \caption{Swooping chassis trajectory stages.
    (a) shows starting stage; chassis has negative attitude due to lowering hind legs in (b), reaching the lowest point of concave-up trajectory in (c) and (d), and raising the cutting appendage in the final portion of concave-up trajectory in (e).}
    \label{fig:swoop}
\end{figure}

We divided the concave-up trajectory motion into five stages (Figures \ref{fig:swoopA}-\ref{fig:swoopE}) for understanding the leg pair motion in each stage.
The BigANT began in the `slack' stage where all the legs were near the apex of their trajectories.
The hind leg pair then reached the lowest point of the toe trajectory, putting the chassis at a negative attitude to the horizontal.
The front and hind pair moved simultaneously in opposite directions to make the cutting appendage attachment point reach the lowest point of the chassis trajectory.
The mid legs were swiftly lowered while the hind legs pushed slightly forward to complete the `swooping' motion. 

We determined this picking motion ad-hoc, however, given that the robot has sufficient actuation to control fore-and-aft motion, height above ground, and pitch angle simultaneously, it is likely straightforward to design a picking motion trajectory using some form of inverse kinematics computation. 

\section{Computer Vision Based Method for Detecting Dandelions}
We used an Intel RealSense L515 RGB-D camera for detecting dandelions to support motion planning.
We mounted the camera at a distance of 0.20m from the vertex of the cutting appendage on the chassis and we accounted for this in the `distance to dandelion' parameter in our code, so that BigANT would not continue walking past the dandelion.
We processed the color (Red-Green-Blue) and depth streams concurrently and we obtained them in the same resolution (640x480) to avoid frame misalignment issues.

For simplicity, we considered a `dandelion' to be any yellow colored globular blob in the field of view.
This also allowed us to test our code with a small inflated yellow balloon and focus  our attention on BigANT's robotic behaviors, while setting us a computer vision framework which future schemes could extend.

\begin{figure}[!htb]
    \centering
    \includegraphics[width=3.1in]{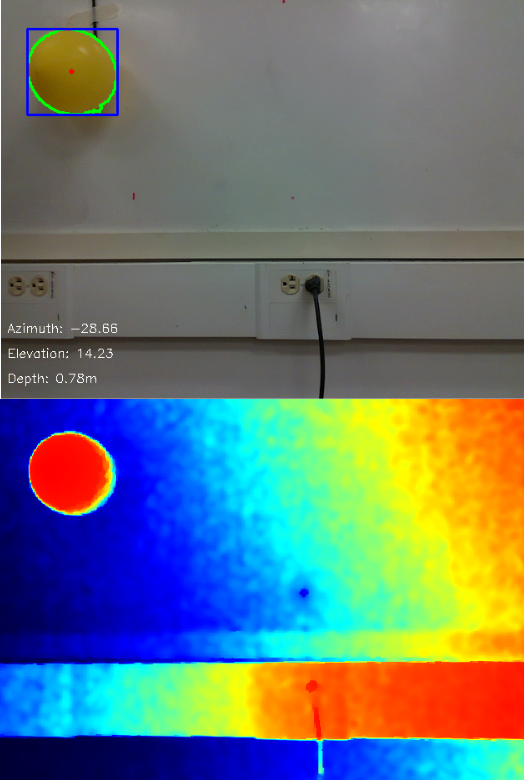}
    \caption{RGB (top) and depth frames show how the dandelion's azimuth, elevation, and depth are tracked; the red dot represents the centroid, the bounding box is in blue, and the contour is green.}
    \label{fig:RGBD}
\end{figure}

Our code used the \texttt{OpenCV 4.5.2.54} and \texttt{NumPy 1.21.0} Python libraries.
We identified the `dandelion' using hue and saturation thresholding in a hue-saturation-value (HSV) representation of the image (Figure \ref{fig:RGBD}).
We used moments of inertia along the vertical and horizontal axes to identify the geometric center of the `dandelion'.
To distinguish between yellow spherical objects and yellow non-spherical objects, we defined a circularity variable as $c = 4\pi A/P^2$, where $A$ and $P$ denote area and perimeter of the object respectively.
When circularity $c$ was in the range $0.4 < c < 1.45$, the yellow object satisfied the condition for being a `dandelion'.

\begin{figure}[!htb]
    \centering
    \includegraphics[width=2.1in]{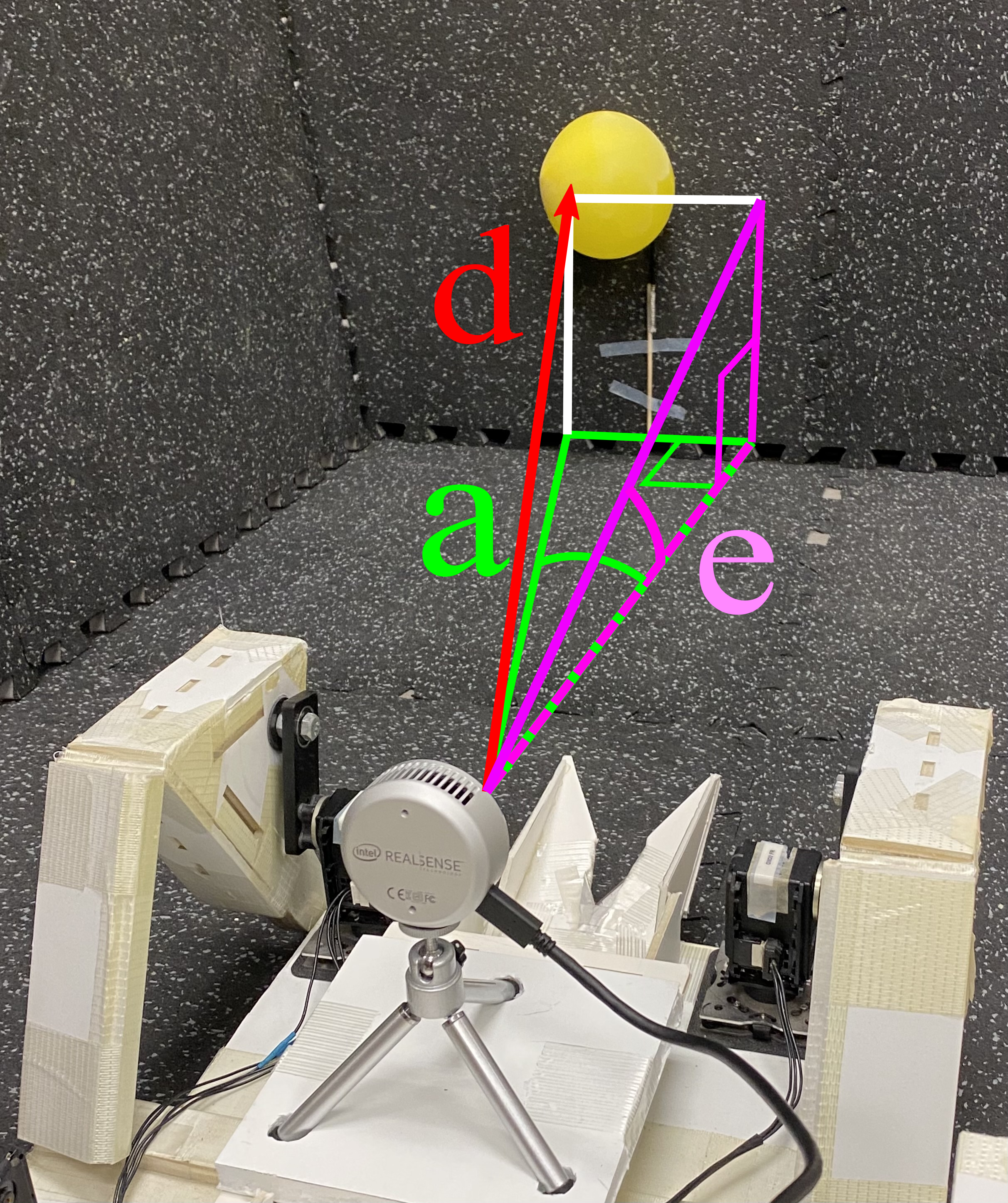}
    \caption{How azimuth, elevation, and distance ($a$, $e$, $d$) are defined.}
    \label{fig:AED}
\end{figure}

We reduced the location of the `dandelion'  to azimuth, elevation, and distance (Figures \ref{fig:RGBD} and \ref{fig:AED}) using the centroid pixel location and the fact that the frame had a field of view of 54\degree {} (horizontal) by 40\degree {} (vertical).
We obtained the remaining distance to the dandelion from the depth frame by taking the mean of depth values contained in the  bounding box obtained from the RGB frame. 

Azimuth, elevation, and distance (referred to as ($a$, $e$, $d$) for concision) can be used to express the vector from BigANT's onboard camera position to the dandelion in spherical coordinates.
It can be shown that the ($a$, $e$, $d$) coordinates translate to ($r$, $z$, $\theta$) as $r = d$cos($\gamma)$, $z = d$sin($\gamma)$, and $\theta = a$, where $\gamma = \arctan(1/(\cos(a)\tan(e)))$ (Figure \ref{fig:AED}).
For the purpose of navigating toward the dandelion, we used azimuth $a$ and distance $d$; elevation played a role in determining whether the dandelion was within picking scope of the cutting appendage or not.

\section{Algorithmic Control Scheme for Motion Planning}
\begin{figure}[!htb]
    \centering
    \subfigure[]{
    \label{fig:schem1A}
    \includegraphics[width=1.25in]{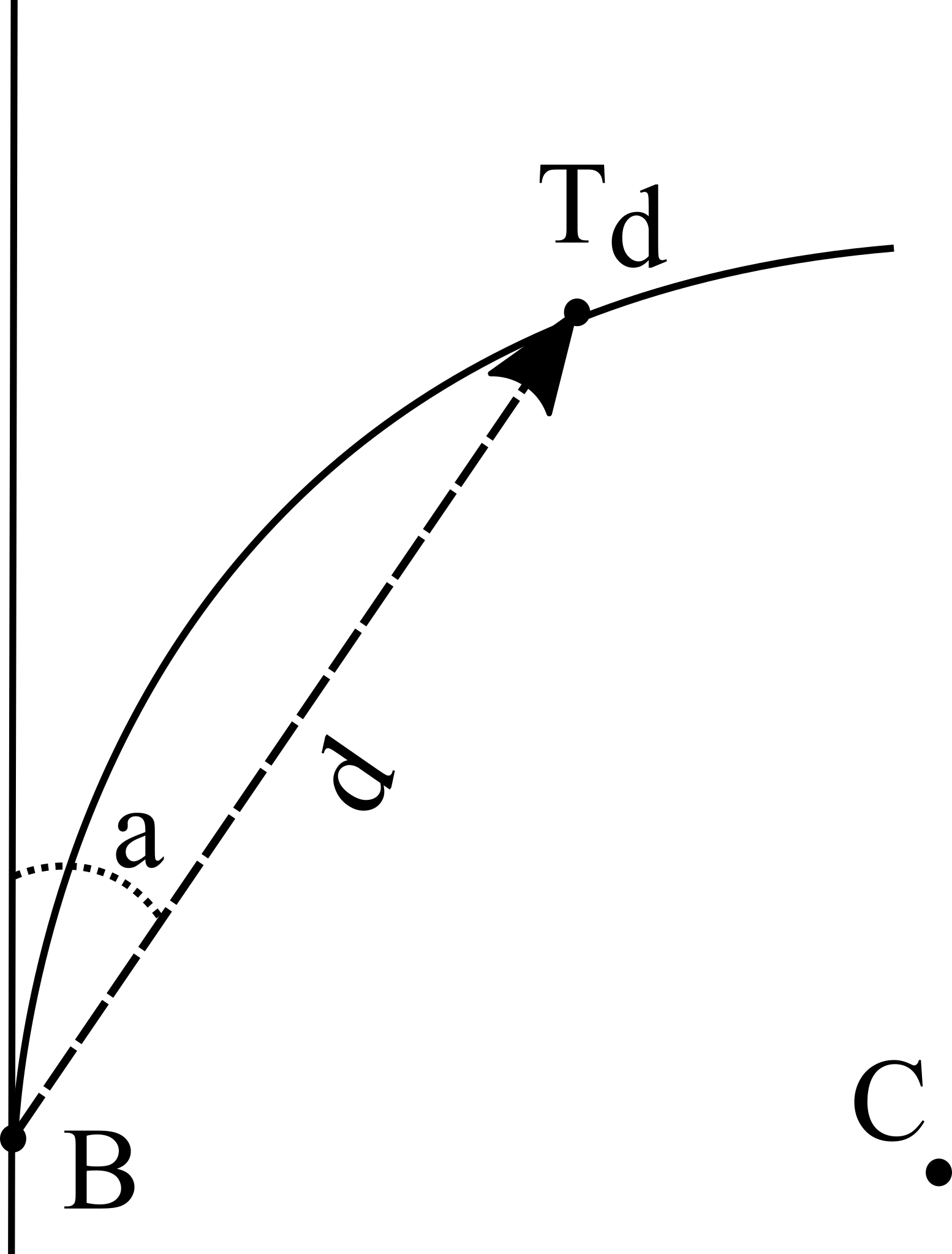}
    }
    \subfigure[]{
    \label{fig:schem1B}
    \includegraphics[width=2.0in]{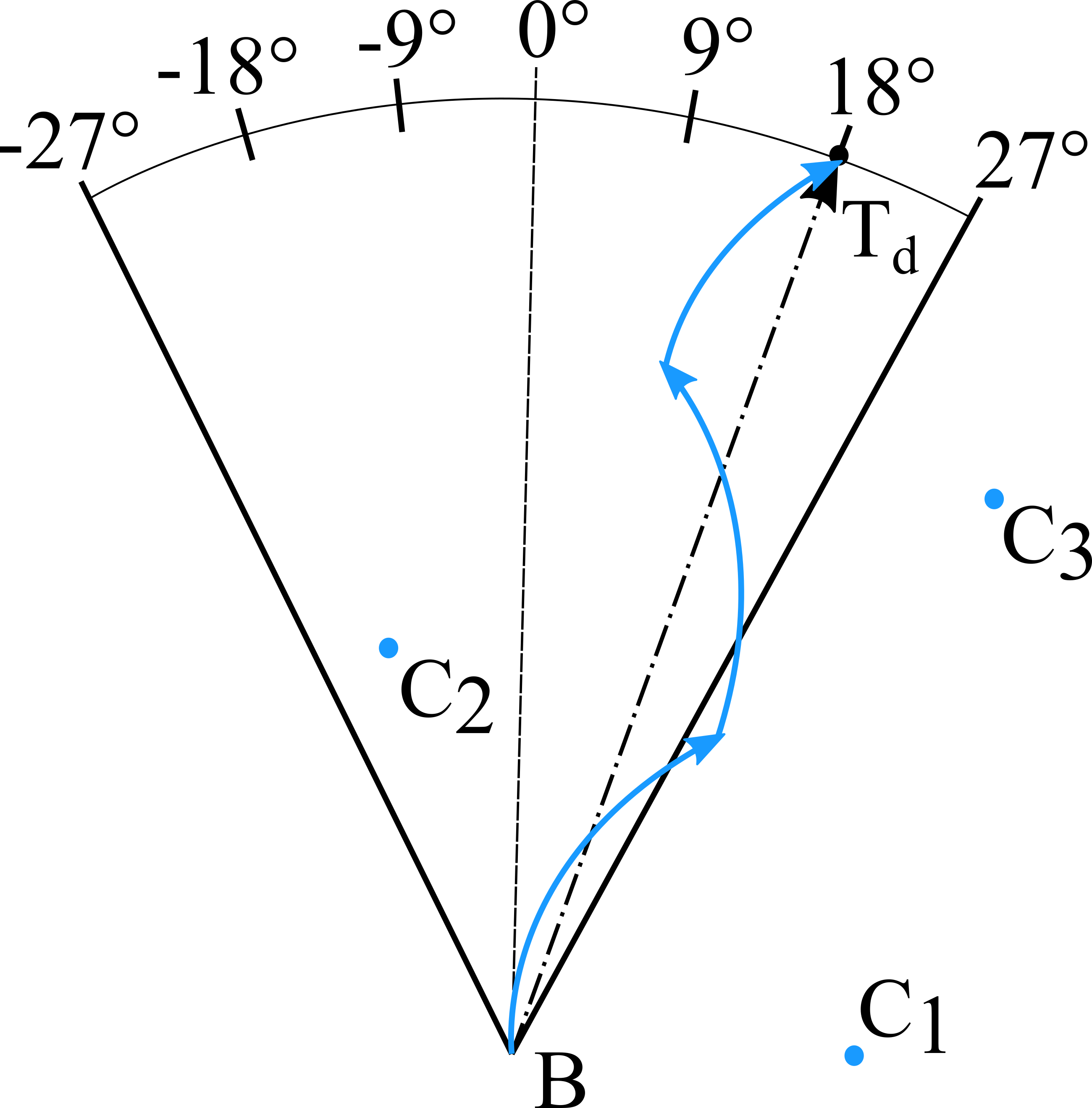}
    }
    \caption{(a) Given azimuth $a$ and distance $d$, a vector in polar coordinates can be defined from BigANT's cutting appendage $B$ and dandelion's position $T_{d}$. An arc with center $C$ resembles BigANT's steering path. (b) shows an example of successive steering arc paths shown to reach the target dandelion placed at 18\degree azimuth and 1.50 m distance. The centers of the arcs which overshoot the target are $C_1, C_2, \and, C_3$.}
    \label{fig:schem1}
\end{figure}

The two parameters we used to formulate the path to the dandelion were azimuth and distance.
With steering (based on definition of turning versus steering in \cite{zhao2020multi}) involved, BigANT's path resembled an arc on which two points represented BigANT's initial position and the target, i.e. the dandelion's position (Figure \ref{fig:schem1A}). 
For instructing BigANT to steer for a specified distance, we analyzed two approaches.
The first approach entailed dead reckoning by walking for a time period calculated from a measured stride length and the gait frequency.

For example, for a gait frequency of $f = 0.16$ Hz, the overall gait cycle time period was $T = 1/f = 6.25$ s.
We experimentally measured that approximately $3$ cycles were needed to walk $1.00$ m, allowing us to estimate $18.75 d$ seconds of walking to go $d$ meters.
The drawback of this method was that the number of cycles to cover a certain distance was not always the same due to variability in the walking gait.
Nevertheless, this naive approach was useful for initial prototyping.

We replaced this scheme with a feedback based policy using the parametric steering gait available for the BigANT \cite{zhao2020multi}.

Here too we considered a dead-reckoning solution first.
We placed a dandelion at several azimuth points along an arc which was 1.50 m from BigANT's cutting appendage (Figure \ref{fig:schem1B}) and explored turn values manually according to whether the BigANT overshot or undershot the dandelion.
Overshooting in this context refers to BigANT steering past the dandelion's azimuth such that the steering direction has to be reversed for it to reduce the difference angle between the cutting appendage and dandelion azimuth.
We produce a calibration curve of turn value as a function of azimuth such that BigANT slightly overshot the dandelion while steering. 
This allowed us to produce successive overshooting paths, which eventually led to the dandelion (Figure \ref{fig:schem1B}) by BigANT autonomously choosing appropriate turn values.

We considered a second steering control scheme which relied on varying a steering parameter $s$ in the shaft angle equations (3) and (4) to modify the phase of the middle legs in an anti-symmetric manner \cite{zhao2020multi}. 
With azimuth and distance, we can define a target point like before, and identify an arc (Figure \ref{fig:AED} (right) in \cite{zhao2020multi}) that contains the start and target points.
Solving for the steering input $s$ which corresponds to the arc then gives a turn value $T$ for steering toward the dandelion.
This approach was different because it identified a turn value $T$ using steering arc results generated by varying a steering parameter $s$, whereas the first approach chose $T$ values based solely on the dandelion's azimuth at a distance of 1.50 m.
The strategy involving steering parameter $s$ may give the ability to steer to distances other than 1.50 m.
Both approaches are limited in that steering results produced on one surface would vary from those on another surface due to differences in foot-surface interactions.

\begin{align}
    \psi_{FL} &= \psi_{HL} := b(\varphi)\\
    \psi_{FR} &= \psi_{HR} := b(\varphi + 1/2)\\
    \psi_{ML} &= b(1/2 + \varphi + s k_s \cos(2\pi\varphi))\\
    \psi_{MR} &= b(\varphi - s k_s \cos(2\pi\varphi))
\end{align}   

We used a state machine to control the picking process.
This consisted of the following processes: (1) receive ($a$, $e$, $d$) data to recognize the dandelion's position, (2) set the turn parameter for steering based on the dandelion's azimuth and walk forwards based on the dandelion's distance, (3) inspect dandelion's position with respect to cutting appendage after completing step (2), (4.1) perform swooping motion if dandelion is in the picking zone or (4.2) steer while walking backwards to reapproach missed dandelion, this time by walking in a straight line towards it.
Due to the nature of the distance covering method, where BigANT continuously adjusts its turn parameter for steering, processes (1) and (2) occur concurrently.

We continuously monitored the remaining distance to reach the dandelion and once BigANT was approximately 0.20 m from the dandelion, we stopped walking.
The 0.20 m value was to compensate for the distance between the camera and the cutting appendage's vertex.

If the dandelion's centroid was in a box defined by the azimuthal range -7.0\degree{} to 7.0\degree{} and an elevation range of -15.0\degree {} to 20.0\degree {} of the cutting appendage (the `picking zone') we performed the swooping motion.
We determined that dandelions can be successfully picked within this zone by repeated trials.

When successive overshooting paths failed to move the cutting appendage such that the dandelion was within the picking zone, BigANT steered backwards on a path with the smallest turning radius (turn parameter $|T| = 0.3$) just until the dandelion was visible within the picking zone.
We then approached the dandelion by walking forwards in a straight line path, keeping it in the picking zone.

\section{Conclusions}
To have a net positive economic impact on  agriculture, robots can evolve in one of two ways: become cheap, reliable, and moderately productive, or remain complex and expensive, but be truly spectacularly productive.
Here we focused on the first of these two approaches.

We have shown how the BigANT hexapod robot's capabilities could perform an agricultural pest control task in a lab setting that simulated some features of dandelion picking in an open field.
This illustrates that low-DOF multi-legged robots can perform meaningful field robotics tasks and future work should consider how the robot morphologies can be used to reap the benefits of legs without the cost of leg complexity.
The solution we presented did not involve adding any DoFs to the BigANT or the mechanism (0-DoF cutting appendage) with which it plucks dandelions, illustrating one of the many advantages of legged systems -- their ability to produce novel motions through coordinated motion.

Our work here was more an illustration of a principle than a complete demonstration.
One improvement which would likely be necessary for any field use is the detection of multiple dandelions at the same time, and the ability to track a target dandelion in a field of other dandelions. 
These kinds of computer vision problems are well studied, and we believe existing solutions could be applied.

More interestingly, future work could explore how to control foot placement while performing the dandelion picking task.
One of the great promises of legged robots for agricultural uses is the ability of legs to trample far less of the agricultural fields than wheels do.
Demonstrating how this could be done well with low DOF, low complexity robot morphologies is a promising follow-up to our work here.

\section{Acknowledgements}
Work on this project was funded by the Wang Chu Chien-Wen Summer Research Award.

\bibliographystyle{IEEEtran}
\bibliography{IEEEabrv, DPLRbibliography}

\end{document}